\documentclass[11pt]{article}

\usepackage[final]{acl}
\usepackage{booktabs}

\usepackage{times}
\usepackage{latexsym}

\usepackage[T1]{fontenc}
\usepackage[utf8]{inputenc}

\usepackage{microtype}

\usepackage{inconsolata}

\usepackage{graphicx}

\usepackage{multirow}
\usepackage{booktabs, makecell, array}

\title{When Good OCR Is Not Enough: Benchmarking OCR Robustness for Retrieval-Augmented Generation}

\author{
 \textbf{Lin Sun},
 \textbf{wangdexian},
 \textbf{Jingang Huang},
 \textbf{Linglin Zhang},
 \textbf{Change Jia},
 \textbf{Zhengwei Cheng},
\\
 \textbf{Xiangzheng Zhang}
\\
\\
Beijing Qiyuan Technology
\\
 \small{
   \textbf{Correspondence: Lin Sun} \href{mailto:sunlin1@360.cn}{sunlin1@360.cn}
 }
}

\begin{document}
\maketitle

% \begin{abstract}
% Industrial Retrieval-Augmented Generation (RAG) systems depend on optical character recognition (OCR) to transform visual documents into text. Existing OCR benchmarks rely on character-level metrics, which inadequately measure downstream RAG effectiveness under real-world conditions. We introduce an OCR benchmark for industrial RAG systems covering 11 challenging document types: extreme layouts, high-resolution pages, complex or watermarked backgrounds, historical documents with non-standard reading orders, visually decorated text, and documents containing tables and mathematical formulas. 
% Evaluating recent OCR models with a fixed OCR-first RAG pipeline shows clear performance degradation on realistic industrial documents despite strong conventional benchmark scores. We find that high OCR accuracy does not necessarily ensure strong downstream RAG performance: structural and semantic errors cause retrieval failures despite low WER/CER. Additional analyses show that this mismatch is category-dependent, persists across representative retriever and chunking choices, and is not removed by a simple multimodal generation baseline.
% \end{abstract}

\begin{abstract}
Industrial Retrieval-Augmented Generation (RAG) systems depend on optical character recognition (OCR) to transform visual documents into text. Existing OCR benchmarks rely on character-level metrics, which inadequately measure downstream RAG effectiveness under real-world conditions. We introduce an OCR benchmark for industrial RAG systems covering 11 challenging document types, including extreme layouts, high-resolution pages, complex or watermarked backgrounds, historical documents with non-standard reading orders, visually decorated text, and documents containing tables and mathematical formulas. Evaluating recent SOTA OCR models under a controlled OCR-first RAG pipeline shows clear performance degradation on realistic industrial documents despite strong conventional benchmark scores. We find that high OCR accuracy does not necessarily translate into strong downstream RAG performance: structural and semantic errors can cause substantial retrieval failures even when WER/CER remains low. Further analysis shows that this mismatch is category-dependent, arises through both retrieval-side and downstream generation-side failures, and remains stable across representative OCR-first pipeline choices. The benchmark is publicly available at \url{https://github.com/Qihoo360/InduOCRBench}. 
\end{abstract}

\section{Introduction}
% Retrieval-Augmented Generation (RAG)~\cite{lewis2020rag} has become a cornerstone of industrial document understanding for enterprise QA and knowledge management. Optical character recognition (OCR) serves as the critical entry point that converts visual documents into text, directly bounding downstream retrieval and generation performance. As illustrated in Figure~\ref{fig:fail_example1}, OCR systems often discard visually encoded semantics such as strikethroughs, transforming legally precise contract clauses into apparent gibberish. The downstream LLM then misattributes this induced ambiguity to ``drafting errors'' rather than recognizing missing format cues, highlighting how high character accuracy masks critical semantic loss.

\begin{figure}[t]
\includegraphics[width=\columnwidth]{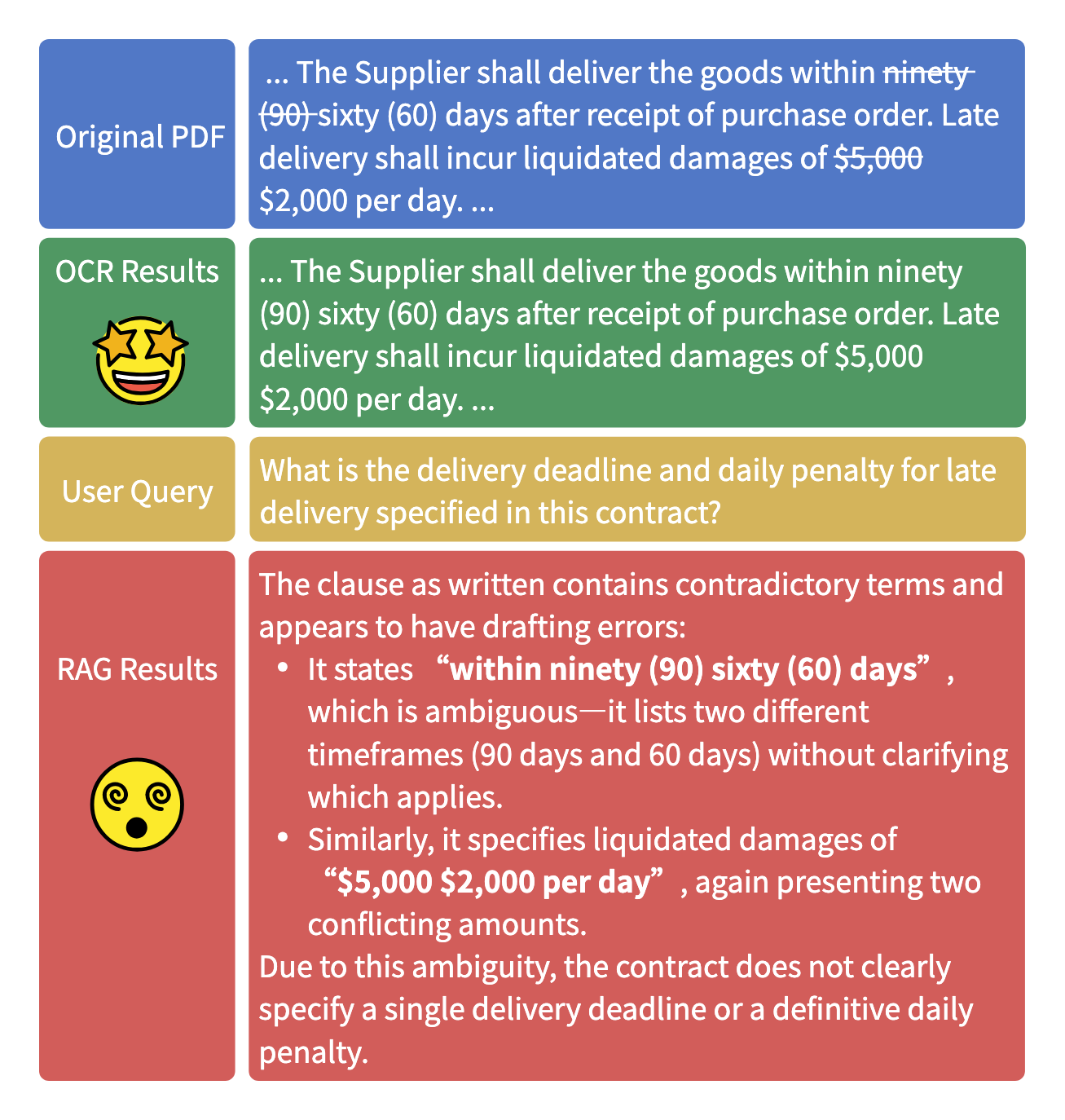}
\caption{OCR strips strikethrough, turning a clear contract into apparent gibberish. LLM blames ``drafting errors'' not missing format, highlighting how high character accuracy hides critical semantic loss.}
\label{fig:fail_example1}
\end{figure}

Retrieval-Augmented Generation (RAG)~\cite{lewis2020rag} has become a cornerstone of industrial document understanding for enterprise QA and knowledge management. Optical character recognition (OCR) serves as the entry point that converts visual documents into text, strongly affecting downstream retrieval and generation performance. As illustrated in Figure~\ref{fig:fail_example1}, OCR systems often discard visually encoded semantics such as strikethroughs, transforming legally precise contract clauses into apparent gibberish. The downstream LLM then misattributes this induced ambiguity to ``drafting errors'' rather than recognizing missing format cues, highlighting how high character accuracy masks critical semantic loss.

\begin{figure*}
    \centering
    \includegraphics[width=1\linewidth]{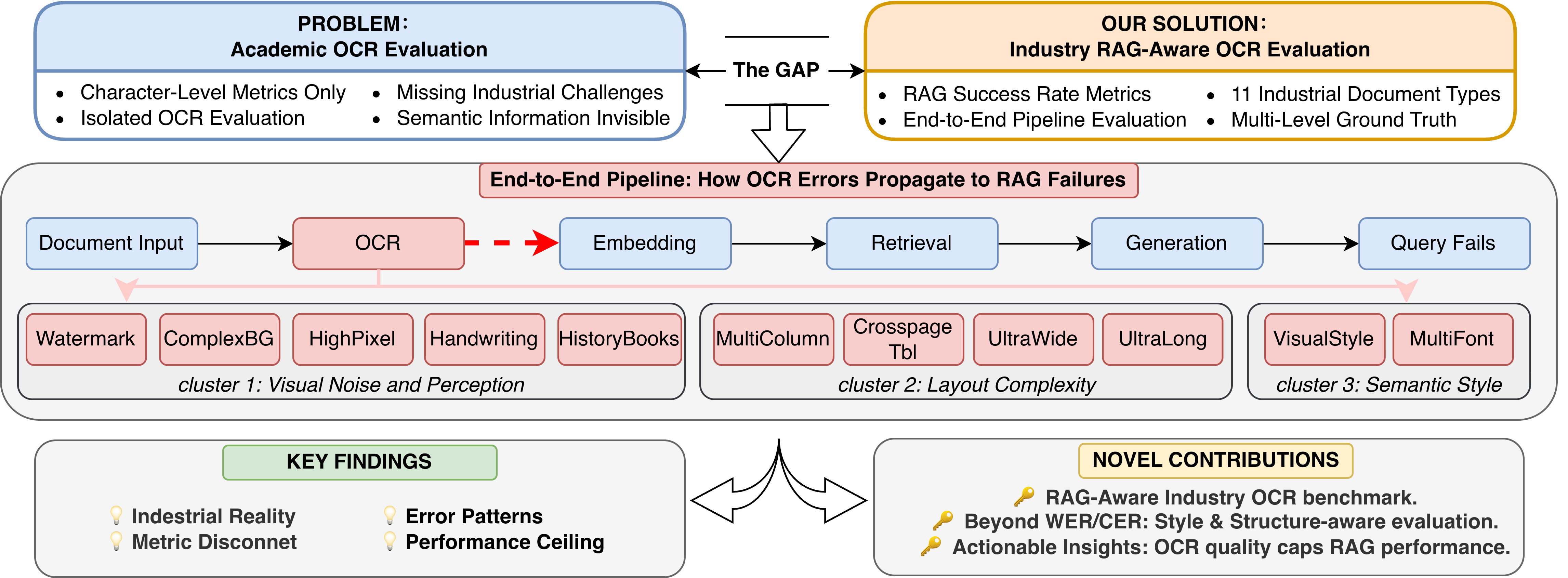}
    \caption{Benchmarking OCR Robustness for RAG.}
    \label{fig:placeholder}
\end{figure*}

Despite progress in OCR technology, evaluation remains dominated by character- and word-level metrics like word error rate (WER) and character error rate (CER). Benchmarks including OCRBench~\cite{liu2024ocrbench} and OmniDocBench~\cite{ouyang2024omnidocbench} assess transcription accuracy in isolation but fail to measure preservation of structural and semantic information essential for downstream retrieval. 
OHRBench~\cite{zhang2025ocrhindersragevaluating} evaluates OCR impact on retrieval through synthetic noise perturbations yet does not model complex industrial characteristics such as extreme page geometries, irregular reading orders, or visually encoded semantics. As structural errors can significantly alter semantics~\cite{anand2024tcocr,abdallah2024tablesurv}, but prior work evaluates prediction accuracy rather than downstream retrieval impact, and RAG evaluation frameworks like RAGAS~\cite{es2024ragas} and ARES~\cite{saadfal2024ares} assume clean textual inputs, overlooking OCR as a critical upstream dependency that fundamentally constrains production retrieval quality.

In practice, industrial RAG systems confront eleven distinct document challenges largely absent from existing benchmarks: extreme layouts (ultra-wide Gantt charts, ultra-long receipts), high-resolution scans with micro-text, complex or watermarked backgrounds, historical documents with non-standard reading orders, visually decorated text where emphasis encodes semantics, and documents containing tables or mathematical formulas spanning multiple pages. These scenarios expose a mismatch between OCR metrics and RAG requirements: errors appearing minor under WER/CER such as discarding strikethroughs, fragmenting cross-page tables, or misordering multi-column layouts, can substantially affect retrieval despite near-perfect character recognition.

% To bridge this gap, we introduce InduOCRBench, a benchmark for evaluating OCR robustness in industrial RAG systems. Our contributions are fourfold: (1) We construct a benchmark covering eleven real-world document challenge categories frequently observed in industrial workflows but underrepresented in existing evaluations. (2) We systematically evaluate state-of-the-art OCR models across these scenarios, revealing substantial performance degradation despite strong standard benchmark results. (3) We establish an OCR-to-retrieval evaluation protocol demonstrating that high OCR accuracy does not guarantee effective RAG performance, with structural and semantic errors causing disproportionate retrieval failures. (4) Error-type analysis and ablation studies show that within conventional RAG architectures, OCR-induced information loss presents a severe bottleneck that current off-the-shelf components struggle to compensate.

To bridge this gap, we introduce InduOCRBench, a benchmark for evaluating OCR robustness in industrial RAG systems. Our contributions are fourfold: (1) We construct a benchmark covering eleven real-world document challenge categories frequently observed in industrial workflows but underrepresented in existing evaluations. (2) We systematically evaluate recent OCR models across these scenarios, showing clear performance degradation despite strong standard benchmark results. (3) We establish an OCR-to-retrieval evaluation protocol showing that high OCR accuracy does not necessarily guarantee effective downstream RAG performance, with structural and semantic errors causing disproportionate retrieval failures. (4) Error analysis, stage-wise attribution, and robustness studies show that within OCR-first text RAG pipelines, OCR-induced information loss is a strong and stable upstream limiting factor across all RAG architectures.

\begin{table*}[t]
\centering
\small
\caption{InduOCRBench Document Type Taxonomy by Primary Challenge and Impact Severity on RAG Systems.}
\label{tab:induocrbench_taxonomy_industry}
%\begin{tabular}{lp{6cm}p{6cm}}
\begin{tabular}{@{}p{1.6cm}p{6.8cm}p{6.8cm}@{}}
\toprule
\multicolumn{1}{c}{\textbf{Category}} &
\multicolumn{1}{c}{\textbf{Technical Challenges}} &
\multicolumn{1}{c}{\textbf{Impact on RAG}} \\
\midrule
\multicolumn{3}{l}{\textit{\textbf{Visual Noise and Perception}}} \\
\midrule
\textbf{Watermark} &
Low-opacity and overlapping background text. &
Context pollution; false-positive retrieval; dedup errors \\

\textbf{ComplexBG} &
Low contrast, textures, and gradient interference. &
Recall loss in complex regions; missing evidence \\

\textbf{HighPixel} &
Ultra-HD GPU mem limits; micro-text downsampled. &
Fine-grained facts dropped; numeric reasoning fails \\

\textbf{Handwriting} &
Cursive writing; large inter-writer variation. &
Annotations and marginal evidence lost \\

\textbf{HistoryBooks} &
Vertical layout; traditional characters; degradation. &
Reading order errors; semantic misalignment \\

\midrule
\multicolumn{3}{l}{\textit{\textbf{Layout Complexity}}} \\
\midrule
\textbf{MultiColumn} &
Irregular columns; ambiguous reading order. &
Logical flow corrupted; multi-hop reasoning fails \\

\textbf{CrosspageTbl} &
Table fragmentation across pages. &
Table structure unrecoverable; relational queries fail \\

\textbf{UltraWide} &
Horizontal stretching (e.g., Gantt charts). &
Partial structure loss; global context incomplete \\

\textbf{UltraLong} &
Vertical stretching (receipts, mobile screenshots). &
Long-range dependencies broken \\

\midrule
\multicolumn{3}{l}{\textit{\textbf{Semantic Style}}} \\
\midrule
\textbf{VisualStyle} &
Semantic cues encoded in bold/color/underline. &
Style-dependent semantics lost; intent misinterpreted \\

\textbf{MultiFont} &
Font switching and size variation. &
Structural and emphasis cues ignored \\

\bottomrule
\end{tabular}
\end{table*}

\section{The InduOCRBench}
\label{sec:benchmark}
We introduce InduOCRBench to bridge the gap between traditional OCR metrics and downstream RAG utility in industrial scenarios. Unlike existing datasets focusing on text transcription, InduOCRBench evaluates a model's ability to preserve document structure and visual semantics critical for reasoning in complex business workflows.
\subsection{Construction and Annotation}
\label{subsec:construction}
\paragraph{Stratified Sampling from Industrial Workflows}
% We sampled 10,000 documents from real-world industrial workflows spanning 12 industries and observed a long tail distribution where structurally simple documents dominate while 11 complex categories cause disproportionate RAG failures. Standard random sampling would inflate model scores by over-representing easy cases. We therefore applied stratified sampling focused on the long tail to construct a high-signal evaluation set of 570 documents spanning 3,402 pages, with balanced representation across industries (Education 20.0\%, Government \& NGO 17.7\%, Technology 12.5\%, Healthcare 8.4\%, Finance 6.7\%, and others) and all 11 challenge categories, transforming the benchmark from a general capability test into a diagnostic stress test targeting industrial OCR-RAG bottlenecks (Appendix Figure \ref{fig:doc_domain_dist}).

We sampled 10,000 documents from real-world industrial workflows spanning 12 industries and observed a long-tail distribution where structurally simple documents dominate while 11 complex categories cause disproportionate RAG failures. Standard random sampling would over-represent easy cases, so we applied stratified sampling to construct a high-signal evaluation set of 570 documents spanning 3,402 pages, with balanced representation across industries (Education 20.0\%, Government \& NGO 17.7\%, Technology 12.5\%, Healthcare 8.4\%, Finance 6.7\%, and others) and all 11 challenge categories (Appendix Figure \ref{fig:doc_domain_dist}).

\paragraph{Three Layer RAG Oriented Annotation}
Our annotation schema captures three information layers essential for RAG using Hybrid Markdown Format. Text Content provides standard transcription. Logical Structure employs HTML with \texttt{rowspan} and \texttt{colspan} for complex tables and LaTeX for mathematical formulas to preserve topology and semantics often lost in standard OCR. Visual Attributes annotate formatting cues such as bolding, underlining, and font colors that serve as semantic anchors in RAG scenarios but are typically ignored by conventional metrics.

\begin{table*}
\centering
\caption{OCR Method Performance on OmniDocBench and InduOCRBench.}
\label{tab:ocr_results}
\small
\setlength{\tabcolsep}{4.5pt} % 调整列间距以适配宽度
\begin{tabular}{@{}l *{10}{c}@{}}
\hline
\multirow{2}{*}{\textbf{OCR Method}} &
\multicolumn{5}{c}{\textbf{OmniDocBench}} &
\multicolumn{5}{c}{\textbf{InduOCRBench}} \\
\cmidrule(lr){2-6} \cmidrule(lr){7-11}
& \makecell{Text\\(EDS)$\uparrow$} & \makecell{Formula\\(CDM)$\uparrow$} & \makecell{Table\\(TEDS)$\uparrow$} & \makecell{Read Order\\(EDS)$\uparrow$} & Avg &
Avg & \makecell{Text\\(EDS)$\uparrow$} & \makecell{Formula\\(CDM)$\uparrow$} & \makecell{Table\\(TEDS)$\uparrow$} & \makecell{Read Order\\(EDS)$\uparrow$} \\
\hline

\multicolumn{11}{l}{\textbf{Pipeline Tools}} \\
 PP-StructureV3 & 92.7 & 85.8 & 81.7 & 92.7 & 86.7 & 60.3 & 78.2 & 53.7 & 49.1 & 79.1 \\
 MinerU2        & 79.1 & 76.6 & 70.9  & 77.5 & 75.5 & 66.5 & 80.1 & 63.2 & 56.3 & 81.3 \\
\hline

\multicolumn{11}{l}{\textbf{Close}} \\
Doc2x          & 91.3 & 78.9  & 83.7  & 91.6 & 84.6 & 61.6 & 76.5 & 56.3 & 52.1 & 81.3 \\
\hline

\multicolumn{11}{l}{\textbf{General VLMs}} \\
GPT-4o         & 78.3 & 79.7  & 67.1 & 85.2 & 75.0 & 52.0 & 60.8 & 58.1 & 37.2 & 70.0 \\
Qwen3-VL-235B & 93.1 & 88.1 & 86.2 & 93.2 & 89.2 & 70.9 & 83.3 & 74.8 & 54.6 & 82.1 \\
Gemini-2.5 Pro & 92.5 & 85.8 & 85.7 & 90.3 & 88.0 & 74.5 & 83.1 & 77.2 & 63.3 & 81.1 \\
\hline

\multicolumn{11}{l}{\textbf{Specialized VLMs}} \\
Deepseek-OCR   & 92.7 & 83.4 & 85.0 & 91.4 & 87.0 & 61.5 & 75.5 & 61.8 & 47.1 & 81.8 \\
Hunyuan-OCR    & 95.8 & 94.7 & 91.8 & 94.3 & 94.1  & 68.1 & 86.1 & 65.6 & 52.5 & 85.7 \\
MinerU2.5      & 95.3 & 88.5 & 88.2 & 95.6 & 90.7 & 72.5  & 81.8 & 75.4 & 60.3 & 84.4 \\
PaddleOCR-VL   & 96.5 & 91.2 & 90.9 & 95.7 & 92.9 & 78.2 & 88.1 & 74.6 & 72.0 & 85.6 \\
\hline
\end{tabular}
\end{table*}

%\paragraph{Rigorous Quality Assurance}
\paragraph{Quality Control}
A three stage Human in the loop pipeline ensures annotation reliability: Annotator Self Correction, Cross Validation, and Stratified Sampling Audit with a 98\% accuracy threshold. Workflow statistics show 66\% of samples required 1--2 revision rounds to meet structural consistency standards, underscoring the task difficulty compared to standard OCR benchmarks (Appendix \ref{app:quality_control}).

% A three-stage human-in-the-loop pipeline supports annotation reliability: Annotator Self Correction, Cross Validation, and Stratified Sampling Audit with a 98\% accuracy threshold. Workflow statistics show 66\% of samples required 1--2 revision rounds to meet structural consistency standards (Appendix \ref{app:quality_control}).

\subsection{The InduOCRBench Taxonomy}
\label{subsec:taxonomy}
We classify the 570 documents in InduOCRBench into 11 distinct types organized into three clusters: visual perception, layout complexity, and semantic style (Table \ref{tab:induocrbench_taxonomy_industry}). Each type receives an annotation reflecting how OCR errors directly break retrieval or reasoning rather than merely degrading recognition accuracy. This taxonomy enables structured failure analysis that explicitly links recognition errors to downstream RAG breakdowns (Appendix \ref{app:taxonomy}).

\section{Experiments}
\label{sec:experiments}
We evaluate SOTA OCR engines on RAG to investigate correlation between OCR quality and RAG accuracy, testing if perfect WER implies retrieval.
% We evaluate recent OCR engines on RAG to examine how OCR quality relates to downstream RAG utility.

\subsection{Experimental Setup}
\paragraph{Baselines and Implementation} We benchmark 10 OCR models across four paradigms: pipeline tools (\textit{PP-StructureV3}~\cite{paddleocr2025}, \textit{MinerU2}~\cite{wang2024mineru}) representing industrial standards; general VLMs (\textit{GPT-4o}~\cite{openai2024gpt4ocard}, \textit{Qwen3-VL-235B-A22B-Instruct}~\cite{Bai2025Qwen3VLTR}, \textit{Gemini-2.5 Pro}~\cite{comanici2025gemini25pushingfrontier}) capable of document understanding without OCR specialization; specialized OCR-VLMs (\textit{DeepSeek-OCR}~\cite{wei2025deepseek}, \textit{HunyuanOCR}~\cite{hunyuanvisionteam2025hunyuanocrtechnicalreport}, \textit{MinerU2.5}~\cite{niu2025mineru25decoupledvisionlanguagemodel}, \textit{PaddleOCR-VL}~\cite{cui2025paddleocrvlboostingmultilingualdocument}) fine-tuned for text-rich scenarios; and closed commercial solution \textit{Doc2X}\footnote{https://doc2x.noedgeai.com/} providing proprietary parsing. General VLMs received standardized prompts for structured Markdown output, pipeline tools used default configurations, and closed solutions accessed via official APIs.
All OCR outputs are evaluated using a unified RAG pipeline to isolate the effect of recognition quality. Full configurations are provided in Appendix \ref{app:rag_setting}. RAG performance is evaluated via RAGAS framework using \textit{Answer Accuracy} (correctness of generated answers) and \textit{Context Recall} (evidence coverage in retrieved passages), both scored by GPT-OSS-120B~\cite{openai2025gptoss120bgptoss20bmodel}. Unless specified, RAG accuracy refers to Answer Accuracy.
We additionally report compact analyses varying retriever, chunking strategy, and generator modality to test whether the main pattern depends on a single downstream configuration.

\subsection{Robustness Gap in Industrial Scenarios}
\label{subsec:ocr_perf}

\paragraph{Distribution Shift Induces Systematic Performance Regression}
Models achieving near perfect scores on OmniDocBench decline sharply on InduOCRBench. PP-StructureV3 drops from 86.7\% to 60.3\% (26.4 points), GPT 4o from 75\% to 52\%, and PaddleOCR-VL from 92.9\% to 78.2\%. This universal regression across architectures confirms existing benchmarks fail to capture real world industrial document distributions. High Normal subset scores (e.g., 88.5\% for MinerU2.5) validate models handle standard data competently, proving the gap stems from complex industrial characteristics rather than inherent incapacity.

\paragraph{Structural Elements Drive Disproportionate Error Rates}
% Text recognition (EDS) remains relatively robust while table (TEDS) and formula (CDM) metrics deteriorate severely. Hunyuan-OCR maintains 86.1\% text accuracy on InduOCRBench but table recognition collapses from 91.8\% on OmniDocBench to 52.5\%. GPT-4o exhibits even worse table understanding at 37.2\%. This structural fragility critically impacts RAG systems where industrial document semantics depend on precise tabular and mathematical alignment under represented in conventional benchmarks.
Text recognition (EDS) remains relatively robust while table (TEDS) and formula (CDM) metrics deteriorate more severely. Hunyuan-OCR maintains 86.1\% text accuracy on InduOCRBench but table recognition drops from 91.8\% on OmniDocBench to 52.5\%. GPT-4o shows even weaker table understanding at 37.2\%. This pattern is especially relevant for industrial documents whose semantics depend on precise tabular and mathematical alignment.

\paragraph{Extreme Layouts Expose Architecture Dependent Vulnerabilities}
% UltraLong and UltraWide categories devastate most systems: GPT-4o scores merely 2.8\% and 3.3\%, while specialized VLMs like PaddleOCR-VL show better resilience at 42.1\% and 63.4\% yet still lag far behind normal document performance. HistoryBooks reveals stark architectural divergence: MinerU2 fail completely at 0.1\% due to non standard reading orders, whereas VLMs (Qwen3-VL-235B) adapt significantly better at 87.1\%. These results underscore industrial processing requires robustness against visual anomalies such as crosspage tables and watermarks smoothed out in standard benchmarks.
UltraLong and UltraWide remain challenging for most systems: GPT-4o scores 2.8\% and 3.3\%, while specialized VLMs such as PaddleOCR-VL reach 42.1\% and 63.4\% but still trail normal-document performance. HistoryBooks reveals strong architectural divergence: MinerU2 drops to 0.1\% under non-standard reading orders, whereas Qwen3-VL-235B reaches 87.1\%. These results suggest industrial OCR requires robustness to visual anomalies such as extreme layouts, cross-page tables, and watermarks that are underrepresented in standard benchmarks.

\subsection{The OCR-RAG Disconnect}
\label{subsec:wer_illusion}
As shown in Figure \ref{fig:fail_example1}, although OCR achieves perfect character fidelity, it  catastrophically degrades RAG effectiveness by discarding structurally encoded semantics. The original contract unambiguously specifies a 60-day delivery deadline and a \$2,000 daily penalty, with "ninety (90)" and "\$5,000" explicitly invalidated via strikethrough. After OCR strips the formatting, the text becomes "ninety (90) sixty (60) days" and "\$5,000 \$2,000 per day", transforming a legally precise clause into apparent gibberish. The RAG system, operating solely on this corrupted input, fails to identify the effective terms and instead misattributes the induced ambiguity to "drafting errors" in the source document. This demonstrates that conventional OCR metrics measure only lexical preservation while ignoring the retention of semantic carriers such as deletion markup that encodes legal validity, rendering them fundamentally inadequate predictors of downstream RAG accuracy in format-sensitive documents.

% As shown in Figure \ref{fig:fail_example1}, even when OCR preserves most characters, it can still substantially degrade RAG effectiveness by discarding structurally encoded semantics. In the contract example, strikethrough marks invalidate ``ninety (90)'' and ``\$5,000'', leaving a 60-day deadline and a \$2,000 daily penalty as the effective terms. Once OCR removes the formatting, the clause becomes ambiguous and the RAG system misattributes that ambiguity to ``drafting errors'' in the source. This example suggests that conventional OCR metrics capture lexical preservation but not semantic carriers such as deletion markup, making them insufficient predictors of downstream RAG accuracy in format-sensitive documents.

\begin{table*}
\centering
\caption{OCR Method Performance on InduOCRBench. CB(ComplexBackground), HP(HighPixel), UL(UltraLong), MC(MultiColumn), UW(UltraWide), HB(HistoryBooks), HW(Handwriting), MF(MultiFont), VS(VisualStyle), WM(Watermark), CT(CrosspageTable). }
\label{tab:ocr_results_on_induocrbench}
\small
\begin{tabular}{l c c c c c c c c c c c c c}
\hline
\textbf{OCR Method} & \textbf{CB} & \textbf{HP} & \textbf{UL} & \textbf{MC} & \textbf{UW} & \textbf{HB} & \textbf{HW} & \textbf{MF} & \textbf{VS} & \textbf{WM} & \textbf{CT} & \textbf{Normal} & \textbf{Avg} \\
\hline
\multicolumn{14}{l}{\textbf{Pipeline Tools}} \\
PP-StructureV3 & 62.4 & 61.5 & 21.4 & 67.9 & 52.5 & 36.8 & 80.4 & 98.0 & 58.6 & 51.4 & 33.0 & 79.4 & 52.0 \\
MinerU2 & 71.6 & 87.9 & 20.8 & 73.4 & 16.9 & 0.1 & 75.0 & 98.8 & 89.1 & 82.5 & 49.9 & 84.6 & 55.5 \\
\hline
\multicolumn{14}{l}{\textbf{Close}} \\
Doc2x & 76.5 & 77.6 & 6.8 & 67.9 & 8.3 & 4.4 & 77.8 & 99.3 & 86.4 & 85.1 & 32.8 & 81.8 & 51.9 \\
\hline
\multicolumn{14}{l}{\textbf{General VLMs}} \\
GPT-4o & 62.8 & 62.7 & 2.8 & 49.7 & 3.3 & 0.0 & 57.3 & 87.0 & 77.0 & 70.5 & 27.6 & 74.9 & 41.7 \\
Qwen3-VL-235B & 81.2 & 72.9 & 48.2 & 67.8 & 31.4 & 87.1 & 98.1 & 98.0 & 91.0 & 86.8 & 44.5 & 80.0 & 67.2 \\
Gemini-2.5 Pro & 84.2 & 83.2 & 36.9 & 74.5 & 48.2 & 78.9 & 97.0 & 97.0 & 86.8 & 86.2 & 50.6 & 84.9 & 68.6 \\
\hline
\multicolumn{14}{l}{\textbf{Specialized VLMs}} \\
Deepseek-OCR & 80.4 & 77.2 & 5.7 & 64.5 & 5.8 & 26.4 & 87.2 & 98.3 & 81.4 & 83.6 & 30.6 & 83.3 & 53.4 \\
Hunyuan-OCR & 84.0 & 84.1 & 28.0 & 65.5 & 19.6 & 84.7 & 98.4 & 98.1 & 86.8 & 85.1 & 34.9 & 84.3 & 64.1 \\
MinerU2.5 & 87.3 & 91.9 & 23.6 & 77.4 & 31.6 & 39.0 & 72.5 & 98.9 & 87.7 & 90.6 & 53.2 & 88.5 & 62.8 \\
PaddleOCR-VL & 83.2 & 89.2 & 42.1 & 84.4 & 63.4 & 70.5 & 97.5 & 98.6 & 84.3 & 83.6 & 50.3 & 86.9 & 70.6 \\
\hline
\end{tabular}
\end{table*}

\subsection{Error-Type Analysis}
\begin{figure}
    \centering
    \includegraphics[width=1\linewidth]{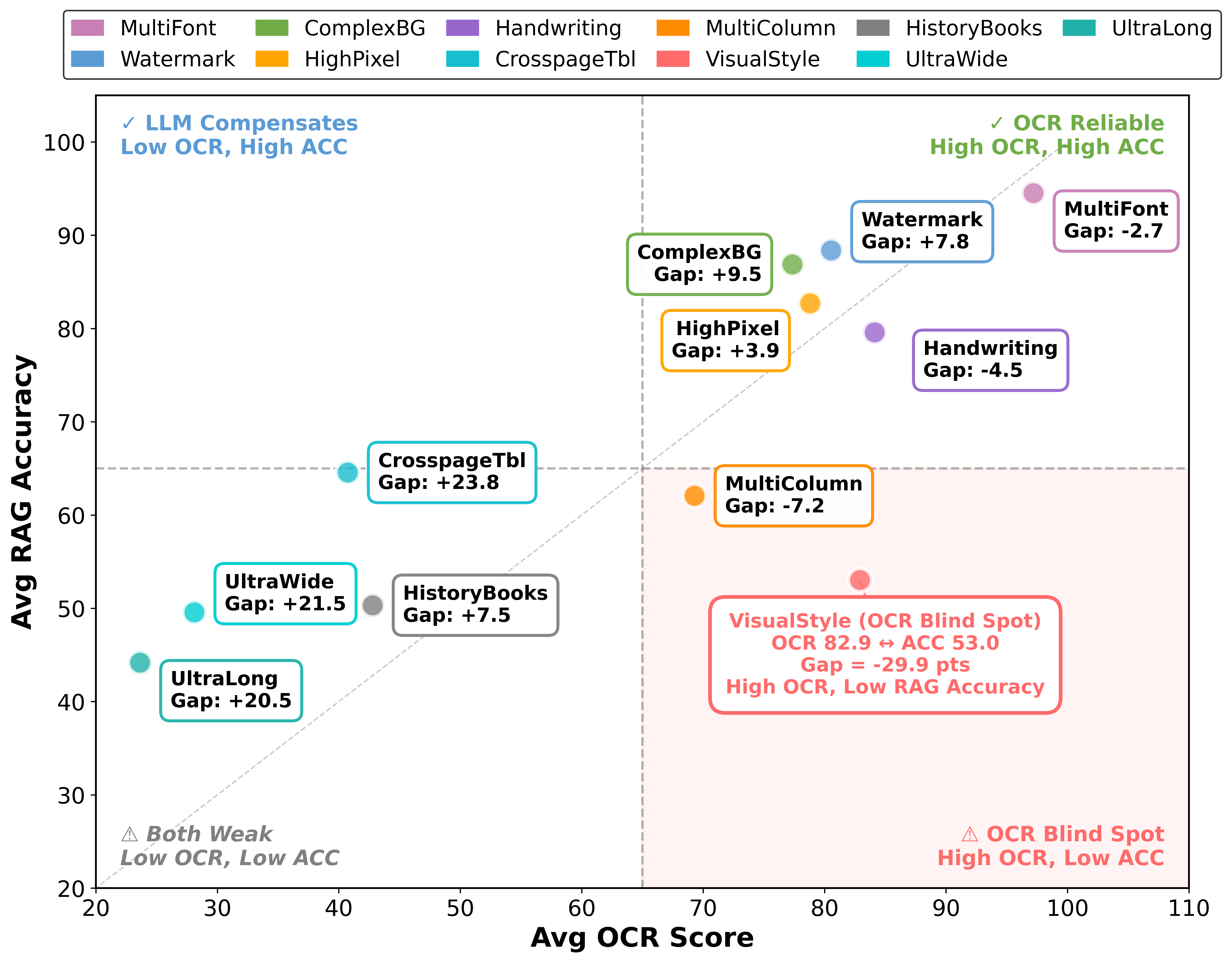}
    \caption{OCR accuracy versus RAG accuracy across document types. Four regimes emerge: OCR Reliable (high-high), LLM Compensates (low-high), Both Weak (low-low), and OCR Blind Spot (high-low). VisualStyle exemplifies the blind spot: 82.9\% OCR accuracy yields only 53.0\% RAG accuracy.}
    \label{fig:ocr_quadrant}
\end{figure}

Figure \ref{fig:ocr_quadrant} partitions document types by OCR and RAG accuracy into four quadrants to identify error patterns governing OCR effectiveness in industrial RAG pipelines. This analysis isolates format semantics loss, geometric fragmentation, contextual compensation boundaries, and dual optimization requirements as dominant failure factors.
\paragraph{Format Semantics Loss Creates Deceptive High Accuracy Failures}
VisualStyle exhibits the largest metric reality gap with 82.9\% OCR accuracy yet only 52.8\% RAG accuracy (30.1\% discrepancy). OCR systems discard visual formatting cues such as strikethroughs and color emphasis encoding critical semantics. MultiFont achieves near perfect alignment at 97.2\% OCR versus 97.5\% RAG since font variations rarely carry standalone semantic meaning. Character level metrics misrepresent extraction quality when format dependent semantics govern document interpretation.
\paragraph{Geometric Fragmentation Induces Cascading Pipeline Failures}
Extreme aspect ratios produce the lowest performance: UltraWide at 28.1\% OCR and 49.1\% RAG, UltraLong at 23.6\% and 42.6\%. CrosspageTbl at 40.7\% OCR and 63.8\% RAG shows partial LLM compensation with RAG exceeding OCR by 23.1 points yet remains below layout intact categories such as ComplexBG at 88.7\% RAG. Geometric fragmentation destroys spatial relationships essential for logical flow reconstruction, and downstream models cannot recover broken relational dependencies. Layout robustness for extreme geometries constitutes a foundational requirement.
\paragraph{Contextual Redundancy Enables Compensation Only With Structural Coherence}
Watermark at 80.5\% OCR and 90.2\% RAG, ComplexBG at 77.4\% and 88.7\%, and HighPixel at 78.8\% and 85.2\% exhibit LLM compensation where RAG exceeds OCR by 6.4 to 11.4 points. These scenarios preserve global document structure despite localized character corruption, enabling inference through contextual redundancy. HistoryBooks at 42.8\% OCR and 50.2\% RAG demonstrates the compensation boundary where non standard reading orders disrupt logical coherence rather than merely obscuring characters. These results suggest that OCR errors affecting structural coherence are often harder to compensate for than errors affecting character fidelity alone.
\paragraph{Dual Failure Modes Demand Parallel Optimization Priorities}
Quadrant analysis reveals two critical failure modes. Low OCR low RAG failures in UltraWide, UltraLong, and HistoryBooks represent explicit robustness gaps demanding immediate investment in geometric handling and reading order recovery. High OCR low RAG failures in VisualStyle produce silent degradation invisible to conventional monitoring, necessitating architectural evolution toward format aware extraction. The 30.2 point gap in VisualStyle versus 23.1 point compensation in CrosspageTbl demonstrates semantic loss is less recoverable than structural fragmentation. OCR development must simultaneously strengthen geometric robustness and preserve visual semantics as both failure modes independently cripple RAG effectiveness.

\subsection{OCR Fidelity Strongly Affects RAG Performance}

\begin{figure}
    \centering
    \includegraphics[width=1\linewidth]{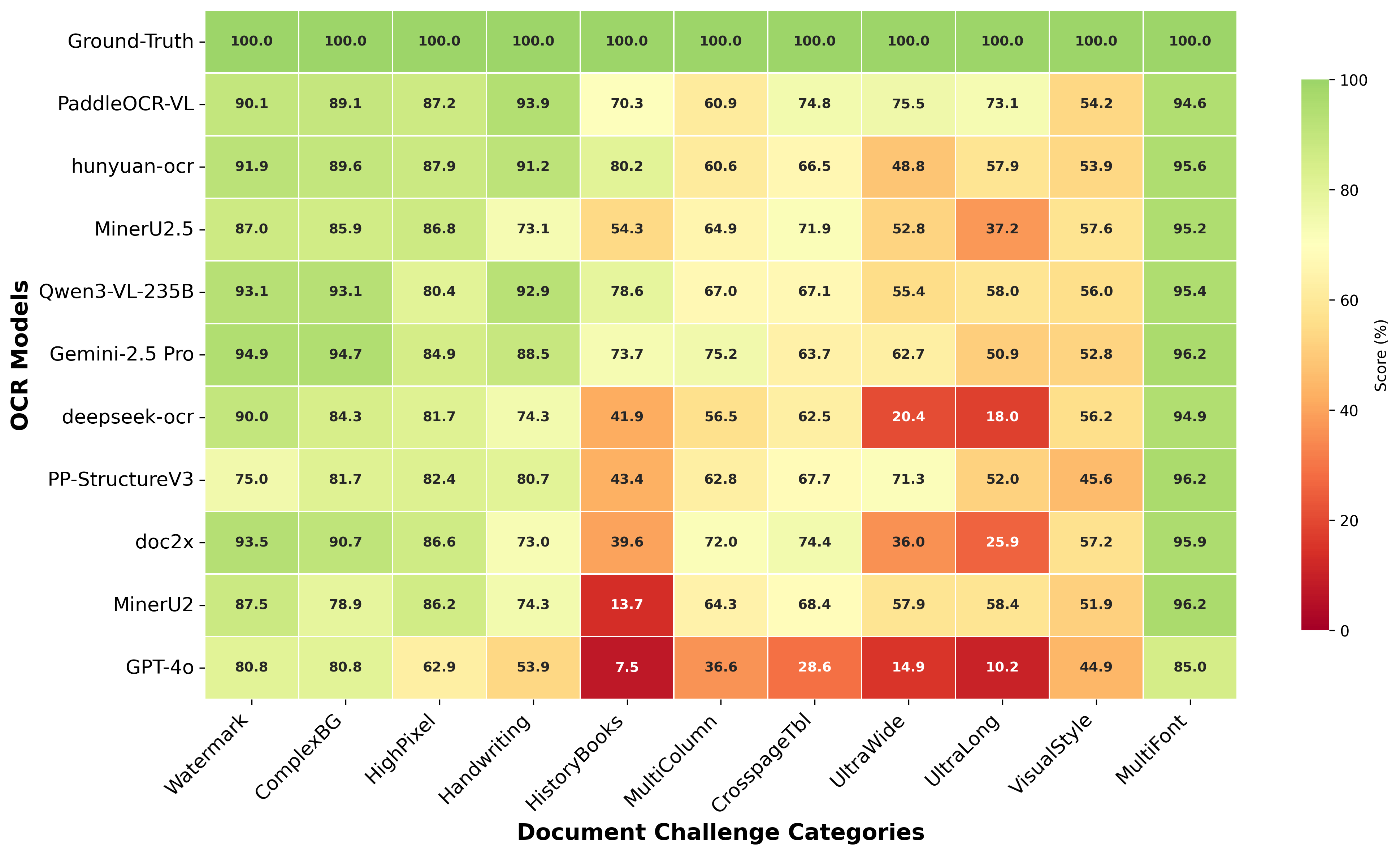}
    \caption{RAG acc (\%) across OCR models and document challenges. Ground-Truth represents perfect OCR. Color indicates performance: green (high) to red (low). }
    \label{fig:heatmap}
\end{figure}

Figure \ref{fig:heatmap} compares RAG accuracy across OCR models against the Ground Truth baseline of perfect extraction. No model exceeds its Ground Truth value across all eleven document types, indicating that under our fixed-pipeline evaluation setup, OCR-induced information loss creates a performance gap that standard RAG components cannot bridge. This suggests that certain structural and semantic losses may be difficult to recover without architecture-level interventions. Ground Truth achieves 100\% RAG accuracy uniformly, representing the empirical upper bound achievable under perfect extraction. Even the strongest models fall short of this bound: Gemini-2.5 Pro attains 97.18\% on Watermark and ComplexBG, PP-StructureV3 reaches 99.63\% on MultiFont. This persistent gap indicates that real-world OCR systems frequently discard information critical for downstream retrieval, resulting in a practical performance limit closely tied to extraction fidelity.

% Figure \ref{fig:heatmap} compares RAG accuracy across OCR models against the Ground Truth baseline of perfect extraction. No model exceeds its Ground Truth value across all eleven document types, indicating that under our fixed OCR-first pipeline, OCR-induced information loss leaves a persistent downstream performance gap. This suggests that certain structural and semantic losses remain difficult to recover even with stronger downstream components. Ground Truth therefore provides an empirical upper bound under perfect extraction, while the remaining gap highlights information frequently lost by real-world OCR systems.

\paragraph{OCR Robustness Is Closely Associated with Achievable RAG Accuracy}
Severe OCR challenges produce larger Ground Truth gaps. UltraLong shows the largest gap, with Ground Truth at 100\% but Qwen3-VL-235B reaching 59.50\%, while UltraWide maintains a 38.22-point gap with Gemini-2.5 Pro at 61.78\%. By contrast, MultiFont remains near ceiling with several models above 98\%. Under identical downstream settings, RAG accuracy varies substantially with OCR quality: HistoryBooks spans 8.63\% (GPT-4o) to 83.45\% (Hunyuan-OCR), UltraLong ranges from 9\% to 74\%, and CrosspageTbl from 25.74\% to 74.59\%. This dispersion suggests that OCR quality strongly shapes the practical ceiling and floor of the evaluated OCR-first pipeline.

\subsection{Additional Analysis}
To localize failure sources, we compare PaddleOCR-VL against Ground Truth using per-category drops in Context Recall and Answer Accuracy, defining $\Delta Recall = Recall_{OCR} - Recall_{GT}$ and $\Delta Acc = Acc_{OCR} - Acc_{GT}$. This reveals three coarse regimes: retrieval-dominated failures, where recall and accuracy drop together (e.g., UltraWide: -21.81\%, -20.16\%; UltraLong: -21.69\%, -21.87\%); generation-sensitive failures, where answer accuracy drops much more than recall (e.g., VisualStyle: -10.56\%, -41.19\%); and compounded failures affecting both stages (e.g., Multi-column: -19.57\%, -31.95\%; CrosspageTbl: -11.10\%, -22.03\%). 

We further test robustness by replacing BGE-M3 with BM25 and Qwen3-8B-Embedding, and HTML-tree chunking with LLM-based semantic chunking (DeepSeek v3.2). The category-wise OCR-induced degradation pattern remains stable, and the average gap changes only slightly under semantic chunking (16.22\% vs. 16.72\%). 

Finally, a simple multimodal baseline that keeps OCR-text retrieval but replaces GPT-5 text generation with GPT-5 vision over retrieved page images does not remove the main effect (73.54\% vs. 52.92\% average accuracy under the same OCR retrieval input). Even with Ground-Truth retrieval text, GPT-5 vision remains below GPT-5 text (53.97\% vs. 89.76\% ). The main exception is VisualStyle(40.32\% → 51.58\%), indicating that visual-semantic cues can help, but even there the GT-OCR gap remains 7.75\%. 

These analyses further support that within OCR-first RAG, OCR fidelity is a strong and stable upstream factor across all RAG architectures.

\label{subsec:ocr_ceiling}

\section{Conclusion and Limitations}
% InduOCRBench establishes that OCR robustness critically constrains industrial RAG performance under realistic document challenges. Our evaluation across eleven failure-prone categories reveals a fundamental disconnect: high character accuracy does not guarantee effective retrieval when structural distortions, layout fragmentation, or loss of visual semantics corrupt downstream inputs. Ablation studies show that within our fixed-pipeline evaluation, stronger embedding models, rerankers, or LLMs demonstrate limited ability to compensate for structural/semantic losses introduced during extraction. Whether specialized error-correction modules could mitigate this gap remains an open question for future work.

% These findings advocate a paradigm shift toward downstream-aware OCR assessment prioritizing structural integrity and semantic fidelity over lexical accuracy alone. We acknowledge three design boundaries: the benchmark's 570-document scale emphasizes diagnostic signal over comprehensiveness; our fixed-pipeline evaluation isolates OCR impact but does not model end-to-end compensation strategies; and we focus on retrieval rather than domain-specific generation tasks to establish a general-purpose foundation. These reflect deliberate trade-offs favoring industrial relevance and diagnostic precision.

InduOCRBench shows that high OCR benchmark scores do not necessarily translate into strong downstream RAG performance on industrial documents. Across eleven failure-prone categories, we observe a consistent mismatch between character-level accuracy and downstream utility when structural distortions, layout fragmentation, or loss of visual semantics affect OCR-first pipelines. Additional analyses further show that this mismatch is category-dependent, can arise through both retrieval-side and downstream generation-side failures, and remains stable across representative retriever and chunking choices. A simple multimodal generation baseline also suggests that the effect is not solely an artifact of using a text-only generator, although categories such as VisualStyle benefit from visual inputs.

These findings motivate downstream-aware OCR assessment that prioritizes structural integrity and semantic fidelity in addition to lexical accuracy. We acknowledge three design boundaries: the benchmark's 570-document scale emphasizes diagnostic signal over comprehensiveness; our evaluation centers on OCR-first pipelines rather than all possible multimodal RAG architectures; and we focus on retrieval rather than domain-specific generation tasks to establish a general-purpose foundation. These reflect deliberate trade-offs favoring industrial relevance and diagnostic precision.

The benchmark dataset and evaluation code are publicly available at \url{https://github.com/Qihoo360/InduOCRBench} to support reproducibility and community extension toward multi-modal grounding and domain-specialized RAG workflows.

% \clearpage

\section*{Ethical Considerations}
InduOCRBench comprises real-world industrial documents collected from enterprise workflows and subsequently de-identified through automated redaction and manual verification. All personally identifiable information and confidential business content were rigorously removed prior to inclusion; the released dataset contains no authentic sensitive data. Document usage complies with fair use provisions for research and benchmarking purposes.

We recognize that robust OCR technologies may be misused for unauthorized document exploitation. Our benchmark exclusively targets improving RAG reliability for legitimate enterprise applications such as contract analysis and knowledge management. The dataset and evaluation protocol are designed solely for research on OCR robustness under industrial conditions.

\section*{Acknowledgements}
We thank the anonymous reviewers and the area chair for their constructive feedback, which helped improve the final version of this paper. We also thank the annotation team for their contributions to dataset construction and quality control.

\bibliography{custom}

\appendix

\section{Detailed Annotation Guidelines}
\label{app:guidelines}

To ensure the high quality of the InduOCRBench and its applicability to downstream RAG tasks, we established a rigorous annotation protocol. We adopted Markdown as the unified format, integrating HTML and LaTeX syntax to resolve common structural fragmentation issues in document parsing and to maximize the restoration of semantic logic.

\subsection{Format Specifications}

\paragraph{Text and Headings} We use standard Markdown syntax for plain text paragraphs. Headings are strictly marked with Markdown headers (\texttt{\#}, \texttt{\#\#}, etc.) to distinguish hierarchical levels and the document outline.

\paragraph{Tables} We reject simplified Markdown table syntax due to its inability to represent complex financial and legal tables. Instead, we standardize on HTML format. This allows us to precisely describe complex table structures, including cell merging (\texttt{rowspan}, \texttt{colspan}), text alignment, and hierarchical header relationships.

\paragraph{Formulas} All mathematical expressions and scientific notations are annotated using LaTeX syntax. This ensures that mathematical symbols and structural relationships are accurately preserved and renderable.

\paragraph{Images} Standard Markdown image reference syntax is used to maintain the position and context of visual elements within the text flow.

\subsection{Handling Specific Document Elements}

\paragraph{Cross-Page Content Merging} For paragraphs split across pages, we perform semantic merging regardless of whether images or other elements are inserted between them. For tables spanning multiple pages, if the original document contains ``Table continued'' or similar indicators, we remove the redundant continuation text and header repetition, merging the data into a single, coherent HTML table object.
    
\paragraph{Hyphenation Handling} For English documents, we specifically address line-break hyphens. If a word is split by a hyphen at the end of a line due to layout constraints, the hyphen is removed, and the word is restored to its complete form to support accurate retrieval.
    
\paragraph{Header and Footer Filtering} In principle, headers and footers are treated as layout noise and removed. However, exceptions are made for content with substantial semantic value, such as data source citations (e.g., ``Data source: Agency X''), which are retained to support source attribution in RAG.
    
\paragraph{Style Retention} We strictly preserve rich text formatting from the original document, including underlining, bolding, font colors, and background colors, converting them into corresponding Markdown or HTML tags.

% \begin{itemize}
%     \item \textbf{Cross-Page Content Merging:}
%     \begin{itemize}
%         \item \textit{Paragraphs:} For paragraphs split across pages, we perform semantic merging regardless of whether images or other elements are inserted between them.
%         \item \textit{Tables:} For tables spanning multiple pages, if the original document contains ``Table continued'' or similar indicators, we remove the redundant continuation text and header repetition, merging the data into a single, coherent HTML table object.
%     \end{itemize}
    
%     \item \textbf{Hyphenation Handling:} For English documents, we specifically address line-break hyphens. If a word is split by a hyphen at the end of a line due to layout constraints, the hyphen is removed, and the word is restored to its complete form to support accurate retrieval.
    
%     \item \textbf{Header and Footer Filtering:} In principle, headers and footers are treated as layout noise and removed. However, exceptions are made for content with substantial semantic value, such as data source citations (e.g., ``Data source: Agency X''), which are retained to support source attribution in RAG.
    
%     \item \textbf{Style Retention:} We strictly preserve rich text formatting from the original document, including \textbf{underlining}, \textbf{bolding}, \textbf{font colors}, and \textbf{background colors}, converting them into corresponding Markdown or HTML tags.
% \end{itemize}

\section{Quality Control Mechanism}
\label{app:quality_control}

To guarantee the precision and consistency of our annotations, we implemented a ``Multi-round Iterative Inspection Mechanism,'' constructing a closed loop for continuous quality convergence.

\subsection{Three-Stage Pipeline}

\paragraph{Stage 1: Annotator Self-Correction.} Annotators start with automated pre-processing results (from machine OCR) and perform item-by-item corrections. This turns the "machine pre-annotation" into a "Personal Reviewed Version", achieving the first round of quality convergence.
    
\paragraph{Stage 2: Cross-Check \& Feedback Loop.} Documents that pass self-correction are assigned to a Quality Inspector (different from the original annotator) for a second round of review.
\begin{itemize}
    \item Inspectors tag specific errors and return the document to the original annotator.
    \item After the annotator fixes the errors, the inspector performs a regression check on the flagged areas.
    \item If issues remain, the document is returned again until all tagged problems are correctly resolved.
\end{itemize}
    
\paragraph{Stage 3: Sampling Audit (The ``Hard Line'').} From the batch of documents that passed Stage 2, we perform random sampling (e.g., 10\%) stratified by document type. These samples are reviewed by Senior Quality Controllers or Project Managers. If the accuracy of the sampled batch falls below 98\%, the entire batch (including unsampled documents) is deemed unqualified and returned to Stage 2 for a full re-review.

% \begin{enumerate}
%     \item \textbf{Stage 1: Annotator Self-Correction.} Annotators start with automated pre-processing results (from machine OCR) and perform item-by-item corrections. This turns the ``machine pre-annotation'' into a ``Personal Reviewed Version,'' achieving the first round of quality convergence.
    
%     \item \textbf{Stage 2: Cross-Check \& Feedback Loop.} Documents that pass self-correction are assigned to a \textbf{Quality Inspector} (different from the original annotator) for a second round of review.
%     \begin{itemize}
%         \item Inspectors tag specific errors and return the document to the original annotator.
%         \item After the annotator fixes the errors, the inspector performs a regression check on the flagged areas.
%         \item If issues remain, the document is returned again until all tagged problems are correctly resolved.
%     \end{itemize}
    
%     \item \textbf{Stage 3: Sampling Audit (The ``Hard Line'').} From the batch of documents that passed Stage 2, we perform random sampling (e.g., 10\%) stratified by document type. These samples are reviewed by \textbf{Senior Quality Controllers} or Project Managers.
%     \begin{itemize}
%         \item \textit{Rejection Policy:} If the accuracy of the sampled batch falls below \textbf{98\%}, the \textbf{entire batch} (including unsampled documents) is deemed unqualified and returned to Stage 2 for a full re-review.
%     \end{itemize}
% \end{enumerate}

\subsection{Quality Statistics}

We recorded the workflow statistics to quantify the rigor of our process. Among the final dataset:
\begin{itemize}
    \item \textbf{33\% (188 documents)} passed inspections on the first attempt without further modification.
    \item \textbf{66\% (376 documents)} required 1--2 iterations of modification and regression testing before meeting the quality standards.
    \item The remaining documents required more than 2 iterations, highlighting the complexity of real-world samples.
\end{itemize}

\section{Detailed Data Taxonomy and Definitions}
\label{app:taxonomy}

Our benchmark comprises 570 documents sourced from real-world business scenarios, covering 11 distinct structural types. All documents are fully annotated. For documents exceeding 20 pages, we applied a truncation strategy, retaining only the first 20 pages to balance structural diversity with annotation efficiency. Notably, while types like ``Handwriting'' and ``History Books'' may originate from open sources, they underwent the same rigorous quality control as proprietary business documents.

The detailed definitions of the 12 document categories are as follows:

\paragraph{Normal} Documents with relatively regular page structures and standardized layout patterns.

\paragraph{UltraWide} Documents where the page width significantly exceeds the height, characterized by horizontal stretching (e.g., Gantt charts, wide spreadsheets).
    
\paragraph{UltraLong} Documents where the page height significantly exceeds the width, characterized by extreme vertical stretching (e.g., mobile screenshots, shopping receipts).
    
 \paragraph{HighPixel} Documents with extremely high resolution and pixel density, imposing higher demands on model parsing precision and GPU memory consumption.
    
\paragraph{ComplexBackground} Documents with complex background structures, such as rich colors, mixed textures, or large-area background images that interfere with text extraction.
    
\paragraph{MultiColumn} Documents with highly complex layouts, featuring multi-column mixed typesetting or cross-column structures (e.g., newspapers, academic papers, magazines).
    
\paragraph{Watermark} Documents containing distinct visible watermarks in the background that overlap with text content.
    
\paragraph{Handwriting} Documents where the entire content consists of handwritten scans (e.g., notes, filled forms).
    
\paragraph{VisualStyle} Documents containing rich semantic formatting, such as underlining, bolding, font colors, or background highlighting, which often denote emphasis or specific meaning.
    
\paragraph{HistoryBooks} Scans of historical ancient books, featuring unique characteristics such as vertical text layout, traditional characters, and woodblock print styles.
    
% \paragraph{MultiFont} Documents containing a mix of different font types within the same page, such as Songti, Fangsong, and others.
\paragraph{MultiFont} Individual documents share a consistent font style, while different documents adopt different font styles, such as Songti, Fangsong, and others.
    
\paragraph{CrosspageTable} Documents containing tables where a single logical table spans across two or more pages, requiring structural merging to restore data continuity.

\label{subsec:doc_domain_dist}
\begin{figure}
    \centering
    \includegraphics[width=1\linewidth]{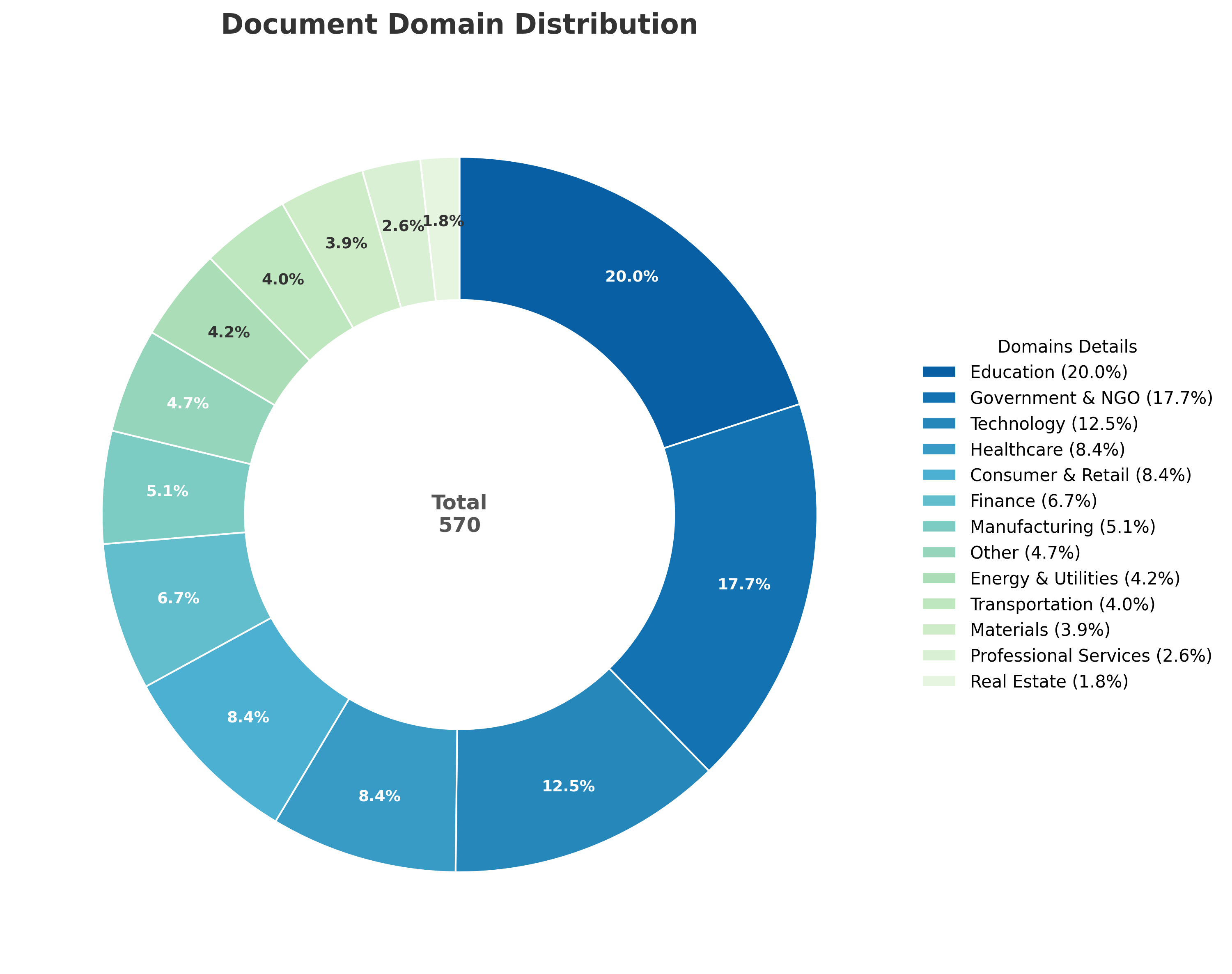}
    \caption{InduOCRBench Document Domain Distribution. }
    \label{fig:doc_domain_dist}
\end{figure}

\begin{figure*}
    \centering
    \includegraphics[width=1\linewidth]{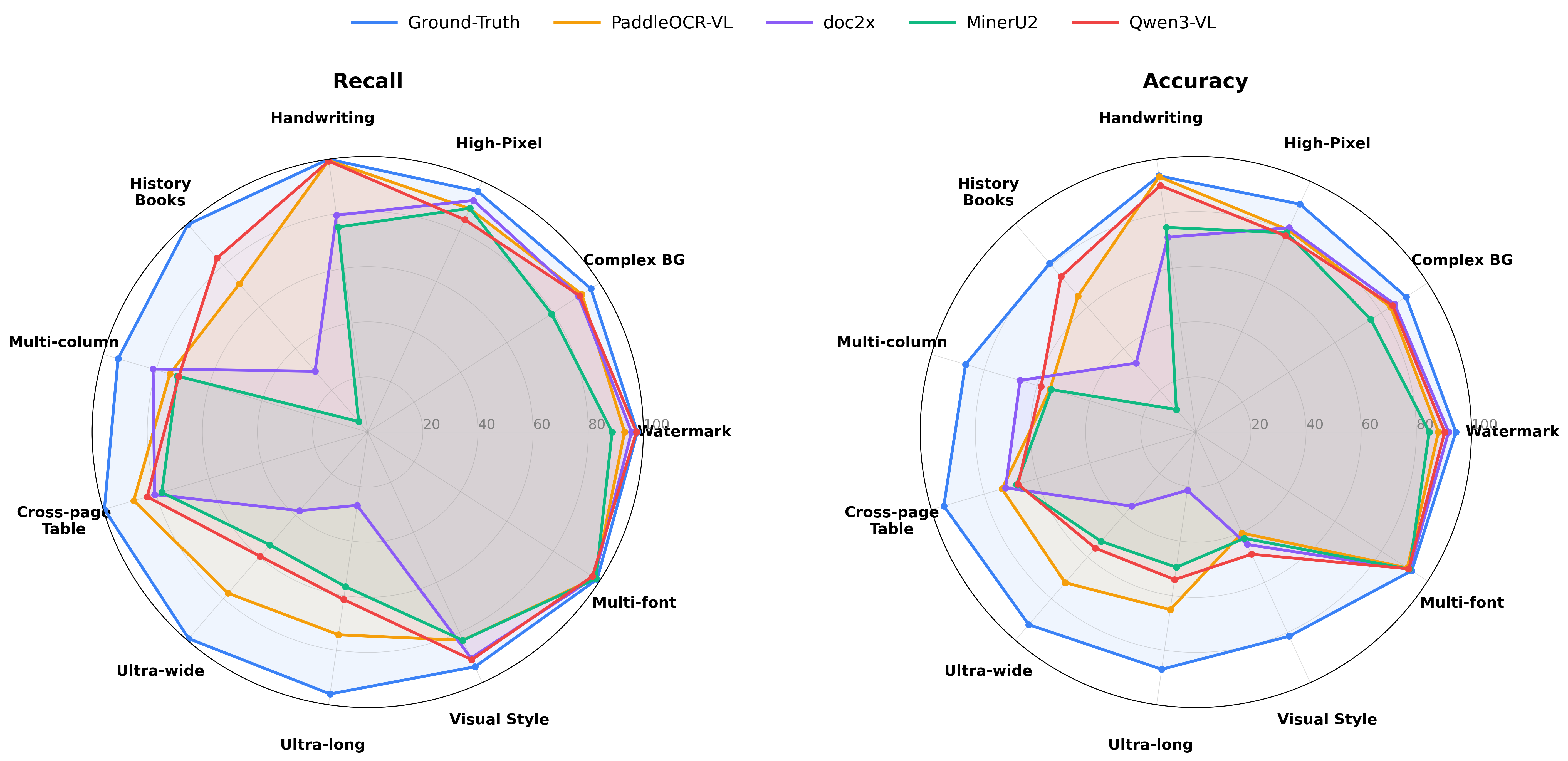}
    \caption{Radar chart comparison of recall (left) and accuracy (right) performance across 11 document challenge categories for five OCR methods. Ground-Truth represents the performance upper bound. Each method shows distinct strengths and weaknesses across different document types.}
    \label{fig:error_type_analyse}
\end{figure*}

\section{RAG Pipeline Details}
\label{app:rag_setting}

\subsection{Query Construction}

For most document types, queries are generated using \texttt{GPT-5} based on predefined categories (e.g., Basic Recognition, Structural Alignment, Cross-Field Continuity, Statistical/Counting, Complex Reasoning). For structurally complex documents such as \textit{CrosspageTbl} and \textit{MultiColumn}, we additionally include structure-sensitive categories (e.g., Structural Alignment Attack, Cross-Page Continuity Attack, Aggregation Attack).

In contrast, the \textit{VisualStyle} category does not use model-generated queries. Instead, its queries are manually crafted to reflect real usage scenarios involving stylistic references (e.g., underlined text, bold text, red annotations). These manually constructed queries are designed to evaluate how OCR errors on visually emphasized regions affect downstream RAG performance.

\subsection{Document Preprocessing and Semantic Chunking}
To preserve structural and semantic integrity during chunking, we first convert the OCR-generated Markdown into HTML. 
The hierarchical structure provided by HTML tags (e.g., \texttt{<h1>}, \texttt{<h2>}, lists, tables, formulas) enables 
reliable grouping of semantically coherent elements. This conversion ensures that logically related units, such as sections, 
tables, and mathematical expressions, remain intact during segmentation. 

We then apply a rule-based segmentation strategy with a maximum chunk length of 256 tokens. Splitting is performed only at 
natural boundaries implied by the HTML tree structure, ensuring that each chunk is both semantically complete and retrieval-efficient.

\subsection{Retrieval-Augmented Generation Pipeline}
We adopt the \textit{Naive} pipeline of the FlashRAG~\cite{FlashRAG} framework to ensure a consistent setup across all OCR systems. 
The retrieval process uses the BGE-M3~\cite{chen2024bge} embedding model to encode document chunks. Dense retrieval is performed using 
a Flat similarity index, from which we retrieve the top-100 candidates. These candidates are subsequently reranked 
by the BGE-Rerank-V2-M3~\cite{li2023making,chen2024bge} cross-encoder to produce the top-10 most relevant passages.

For answer generation, we employ ChatGPT-5. All OCR systems are evaluated with the same retrieval and generation 
settings to isolate the effect of upstream recognition quality on downstream RAG utility.

\subsection{Evaluation Protocol}
We evaluate downstream RAG utility using the Ragas framework, focusing on two metrics:
\begin{itemize}
    \item \textbf{Context Recall}: measures whether retrieved passages contain evidence supporting the ground-truth answer.
    \item \textbf{Answer Accuracy}: evaluates the correctness of the generated answer.
\end{itemize}

Both metrics are computed using the GPT-OSS-120B model as the evaluator to ensure consistent automatic scoring 
across all systems.

\subsection{Additional Implementation Details}
We apply deterministic decoding with temperature set to 0.0 for all generations. All experiments were conducted on a 
cluster equipped with NVIDIA H800 GPUs. The FlashRAG pipeline follows its default setup except for the specified embedding, 
reranking, generation, and indexing components described above.

%\section{Case Study Analysis}
%\label{app:case_study}

\end{document}